\documentclass[letterpaper, 10 pt, conference]{ieeeconf}

\IEEEoverridecommandlockouts                              

\usepackage{graphics} 
\usepackage{graphicx}
\usepackage{subcaption}
\usepackage{multirow}
\usepackage{balance}
\usepackage{tabularray}
\usepackage{array}
\usepackage{multirow}
\usepackage{ragged2e}

\usepackage{amsmath} 
\usepackage{amssymb}  

\usepackage{float}
\usepackage[font=footnotesize]{caption}
\title{\LARGE \bf
Learning from Demonstration Framework for Multi-Robot Systems Using Interaction Keypoints and Soft Actor-Critic Methods 
}

\author{ Vishnunandan L. N. Venkatesh and Byung-Cheol Min% <-this % stops a space
\thanks{This material is based upon work supported by the National Science Foundation under Grant No. IIS-1846221. The authors are with SMART Lab, Department of Computer and Information Technology, Purdue University, West Lafayette, IN 47907, USA
        {\tt\small lvenkate@purdue.edu, minb@purdue.edu }}%
        } 
\begin{document}

\maketitle
\thispagestyle{empty}
\pagestyle{empty}
%%%%%%%%%%%%%%%%%%%%%%%%%%%%%%%%%%%%%%%%%%%%%%%%%%%%%%%%%%%%%%%%%%%%%%%%%%%%%%%%
\begin{abstract}
Learning from Demonstration (LfD) is a promising approach to enable Multi-Robot Systems (MRS) to acquire complex skills and behaviors. However, the intricate interactions and coordination challenges in MRS pose significant hurdles for effective LfD. In this paper, we present a novel LfD framework specifically designed for MRS, which leverages visual demonstrations to capture and learn from robot-robot and robot-object interactions. Our framework introduces the concept of Interaction Keypoints (IKs) to transform the visual demonstrations into a representation that facilitates the inference of various skills necessary for the task. The robots then execute the task using sensorimotor actions and reinforcement learning (RL) policies when required. A key feature of our approach is the ability to handle unseen contact-based skills that emerge during the demonstration. In such cases, RL is employed to learn the skill using a classifier-based reward function, eliminating the need for manual reward engineering and ensuring adaptability to environmental changes. We evaluate our framework across a range of mobile robot tasks, covering both behavior-based and contact-based domains. The results demonstrate the effectiveness of our approach in enabling robots to learn complex multi-robot tasks and behaviors from visual demonstrations. 

\end{abstract}
\section{INTRODUCTION}\label{Intro}

Learning from Demonstration (LfD) represents a pivotal advancement in robotics, shifting paradigmatic approaches towards more intuitive, efficient skill acquisition in intelligent systems. By leveraging human demonstrations, LfD facilitates the teaching of complex behaviors to robots without the need for intricate programming, embodying a natural progression towards more accessible human-robot interactions. This methodology not only simplifies the programming landscape but also heralds a new era of potential applications that span the manufacturing, healthcare, and surveillance sectors \cite{billard2016learning, balakuntala2021learning}. The growing importance of LfD is evident from its increasing presence in academic research \cite{ravichandar2020recent}. 

\begin{figure}[t]
    \centering
    \includegraphics[width=0.47\textwidth]{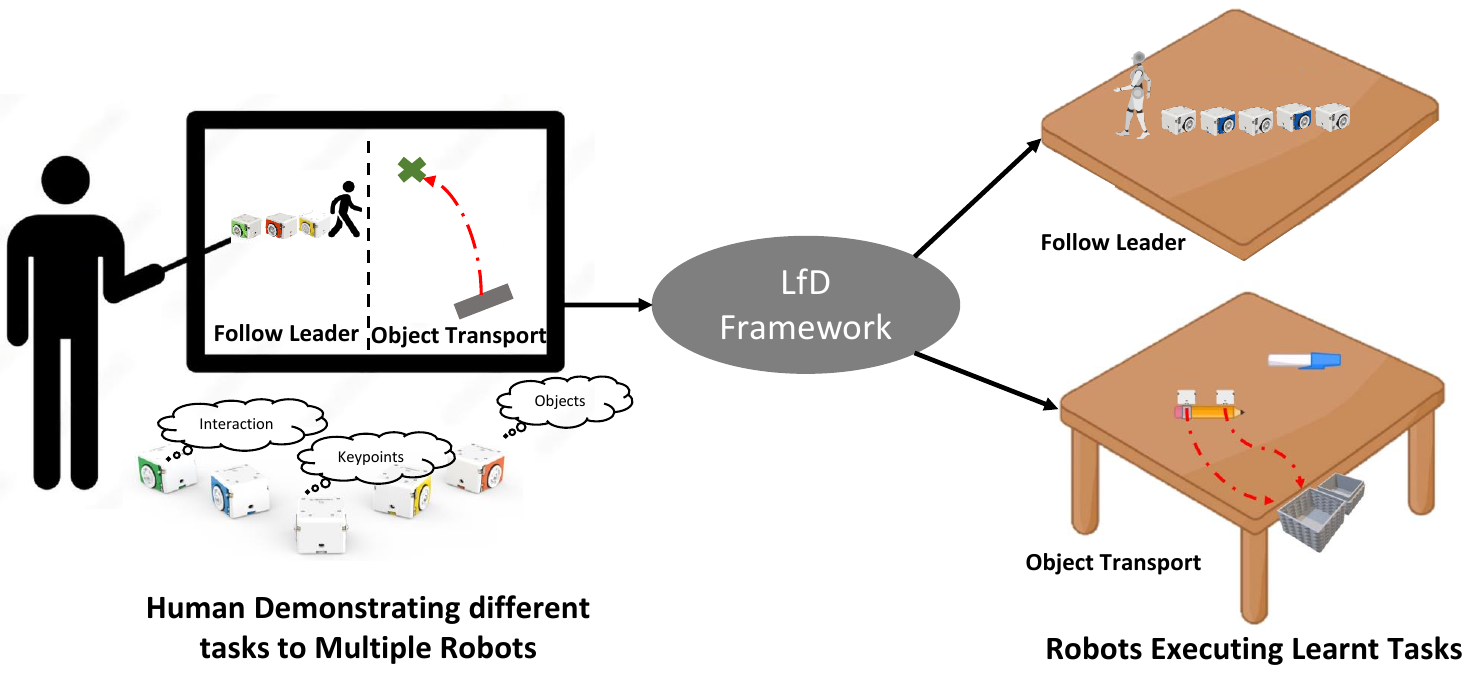}
    \vspace{-5pt}
    \caption{ Concept for  learning from demonstration for multi-robot systems (MRS). The human expert demonstrator shows multiple tasks to the MRS, which are then learned and executed.}
    \label{fig:rob}
    \vspace{-20pt}
\end{figure}
Despite its growing promise, the exploration of LfD within multi-robot systems (MRS) remains nascent, presenting a unique array of challenges and opportunities \cite{argall2009survey,ravichandar2020recent}. The complexity inherent to MRS, marked by intricate robot-robot and robot-environment interactions—significantly compounds the challenges faced in LfD. Multi-robot contexts escalate the variables and potential interactions exponentially, complicating both the design and control of these systems compared to their single-robot counterparts. These complexities necessitate innovative approaches in the development of LfD algorithms capable of navigating the multifaceted dynamics of MRS, including coordination and data processing between multiple agents \cite{ravichandar2020recent}.

Recent research efforts in LfD for MRS have explored a spectrum of demonstration methodologies, ranging from simulator-based to vision-based demonstrations, each with its advantages and limitations. Despite significant advances, current approaches often struggle with task versatility and an over-reliance on extensive demonstration sets. Moreover, the predominant focus on nonvisual demonstration methods hampers intuitive natural human-robot communication, underscoring the critical need for more accessible, vision-based frameworks \cite{huang2017vision}.

Our work introduces a vision-based learning-from-demonstration framework for multi-robot systems, leveraging \textbf{Visual Interaction Keypoints} and \textbf{Soft Actor-Critic (SAC)} methods. The choice of visual cues as the primary mode of demonstration is motivated by their inherent capacity for facilitating intuitive interactions between humans and robots \cite{huang2017vision}, despite their inherent shortcoming in conveying tactile information. To overcome this challenge, we incorporate the SAC algorithm, which enables robots to master tasks that require physical contact. Central to our method is the utilization of a binary decision classifier alongside interaction keypoints, which collectively fine-tune the reward mechanism without resorting to modeled engineering. The interaction keypoints pinpoint crucial moments of interaction within the environment, such as instances of contact between a robot and an object or the proximity of robots to each other, thereby dividing complex tasks into manageable subtasks. This segmentation not only boosts the efficiency of the learning process but also improves the clarity with which MRS can be understood and interpreted. A conceptual overview of our framework is presented in Fig. \ref{fig:rob}. 

MRS encompass various task categories, including navigation and exploration~\cite{burgard2005coordinated}, coordination and communication, decision-making and planning, assembly and manufacturing, manipulation and grasping, and intensive contact-based tasks~\cite{darmanin2017review}. Our framework simplifies this complexity by classifying tasks into \textbf{Behavior-Based} and \textbf{Contact-Based} categories. Behavior-based tasks encompass activities like pattern formation and surveillance, while contact-based tasks involve direct physical interactions, such as pushing or lifting. What distinguishes our framework is its novel use of vision-based demonstrations to effectively learn and execute tasks within these categories. It efficiently processes behavior-based tasks using interaction keypoints from a single clear demonstration. For contact-based tasks, although multiple demonstrations may be required, the approach remains more streamlined and less demanding than traditional machine learning-based methods, enhancing both efficiency and applicability.

The main contributions of this paper are: 
\begin{itemize}
    \item We propose a novel LfD framework for MRS that utilizes \textit{Visual Keypoint} inference and \textit{SAC} methods, addressing the gap in current research. This framework is capable of performing behaviour and contact-based tasks. 
    \item We evaluate our framework through real-time experiments on diverse tasks, demonstrating its efficacy in collaborative behavior and contact-based tasks.
\end{itemize}

The remainder of this paper is organized as follows. We begin with a comprehensive literature review focused on MRS LfD, followed by a detailed definition of the problem. The following sections outline our methodology, experimental setup, and results. We conclude with a discussion of the limitations of our framework and potential directions for future research.

\begin{figure*}[ht]
    \centering
    \includegraphics[width=1\textwidth]{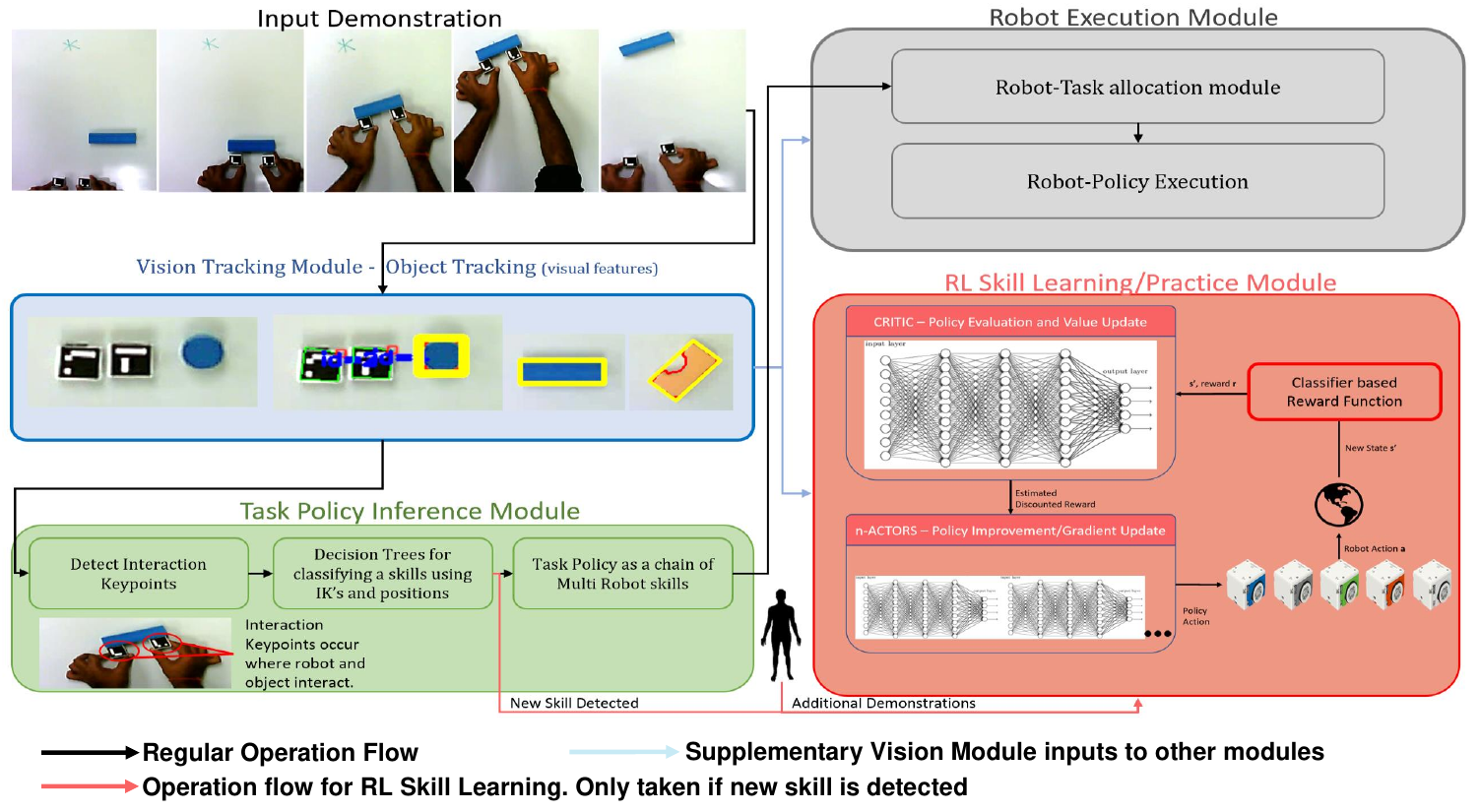}
    \caption{The proposed learning from demonstration for multi-robot systems framework follows a streamlined process: Human experts visually demonstrate tasks, captured by a 2D camera. These demonstrations undergo feature extraction in the \textbf{Vision Tracking Module}. The \textbf{Task Policy Inference Module} segments the demonstrations and identifies \textit{Interaction Keypoints}, forming a \textit{Task Policy}. When new contact skills arise, the \textbf{RL skill Learning/Practice Module}, using SAC networks, learns them with guidance from a binary decision classifier's reward signals. Finally, the \textbf{Robot Execution Module} allocates and executes tasks across multiple robots, showcasing the adaptability of the framework in various environmental conditions.}
    \label{fig:arch}
    \vspace{-12pt}
\end{figure*}

\section{RELATED WORK}

While there is a plethora of research conducted in the domain of single robot learning from demonstration \cite{ravichandar2020recent,argall2009survey,brunke2022safe}, this review of the literature explores MRS in the context of LfD. It covers a range of demonstration methods, from simulator demonstrations to vision-based human demonstrations, emphasizing the significant role of MRS in LfD research.

Notable studies include \cite{chernova2009interactive}, which introduces a confidence execution algorithm for collaborative ball sorting, featuring an adaptive interruption mechanism for when robots require additional human demonstrations due to low confidence. Knepper et al.  discuss task conveyance through geometric CAD designs for the assembly of furniture by robots with specialized roles \cite{knepper2013ikeabot}. Huang et al. combine human demonstrations with vision for bi-manual surgical tasks, employing Gaussian Mixture Models for learning, illustrating collaborative work even within a single robotic system \cite{huang2017vision}. The authors in \cite{martins2010learning} and \cite{sullivan2011hierarchical} demonstrate teleoperation for navigation and door opening and a hierarchical reinforcement learning (RL) framework for abstract behavior tasks, respectively. In robot soccer, Freelan et al. in \cite{freelan2014towards} employs reinforcement learning and state-space automata \cite{simoes2020dataset} to teach set plays, highlighting the domain's extensive research in LfD and reinforcement learning.

However, these advances often face challenges such as task specificity, extensive demonstration requirements, and lack of intuitive vision-based communication. Moreover, the limited research in the context of LfD for MRS shows the inherent complexity when dealing with an MRS that includes interactions that can affect the environment exponentially. Our framework addresses these issues by leveraging vision-based demonstrations and interaction keypoints \cite{balakuntala2019extending,kannan2023upplied} for a wide range of tasks. Inspired by \cite{singh2019end}, we incorporate SAC methods for their real-time efficacy in complex task learning. Crucially, our framework processes behavior-based tasks with Interaction Keypoints from a single, clear demonstration, streamlining the learning process. For contact-based tasks, while multiple demonstrations may still be necessary, our method is significantly more streamlined and less demanding than traditional machine learning approaches \cite{abbeel2004apprenticeship,ravichandar2020recent,nair2018overcoming}, enhancing both efficiency and applicability.

\section{Problem Definition}

Our framework addresses the challenge of instructing MRS to perform tasks based on visual demonstrations, categorized into behavior-based and contact-based tasks. We formalize the inputs, processes, and outputs as follows:

\noindent\textbf{Inputs:}
\begin{itemize}
    \item \( D = \{f_1, f_2, \ldots, f_n\} \): A sequence of \( n \) frames from visual demonstrations captured by a 2D camera.
\item \( O = \{o_1, o_2, \ldots, o_m\} \): A set of \( m \) recognized objects within the frames.
\item \( R = \{r_1, r_2, \ldots, r_k\} \): A set of \( k \) identified robots within the frames.

    \item \( G \): The goal positions for objects and robots that conform to the deduced goal state from the demonstrations.
\end{itemize}

\noindent\textbf{Outputs:}
\begin{itemize}
    \item \( IK = \{ik_1, ik_2, \ldots, ik_p\} \): Interaction Keypoints, indicating significant moments of interaction. \(p\) represents the total number of interaction keypoints identified. 
    \item \( T_P \): A Task Policy for performing the task.
    \item \( S_{RL} \): Skills developed through Reinforcement Learning for contact-based tasks.
\end{itemize}

\noindent\textbf{Problem 1: Behavior-Based Task Learning}.
The behaviour-based task learning problem involves deriving a task policy \( T_P \) from visual demonstrations, utilizing interaction keypoints and the spatial dynamics of robots and objects:
\begin{equation}
\label{eq: tp}
T_P = f(D, IK, O, R, G)
\end{equation}
where the function \( f \) encapsulates the algorithms or set of processes that the LfD framework employs to interpret the visual demonstrations. The resulting task policy \( T_P \) details the actions or behaviors that robots are to perform to complete the demonstrated task.

\noindent\textbf{Problem 2: Contact-Based Task Learning}.
The contact-based task learning problem extends behaviour-based learning by incorporating reinforcement learning for learning skills that involve physical contact interactions in the task, requiring additional demonstrations to train a decision classifier \( C \) for goal state recognition:
\begin{equation}
\label{eq:cont}
S_{RL} = g(T_P, C(D_{add}), IK, O, R, G)
\end{equation}
where the function \( g \) represents the learning process that integrates \( T_P \), \( IK \), \( O \), \( R \), and \( G \), with the output of a decision classifier \( C \) trained on additional demonstrations \( D_{add} \) to synthesize reinforcement learning skills \( S_{RL} \) for contact-based tasks. 

Our aim is to develop an integrated set of execution policies \( \Pi \) that enables the MRS to autonomously perform complex tasks, aligned with the learned behavior-based and contact-based policies, and ultimately matching the goal configuration \( G \):
\begin{equation}
\label{eq:policy}
\Pi = h(T_P, S_{RL}, G)
\end{equation}
where the function \( h \) denotes the integration process that amalgamates the task policy \( T_P \), skills \( S_{RL} \) acquired from reinforcement learning for contact-based interactions, and the goal states \( G \), to produce a set of executable policies \( \Pi \) for the multi-robot system. This integration represents the culmination of the learning from demonstration process, enabling the robots to perform tasks effectively.

This symbolic representation defines the transformation from input demonstrations to executable multi-robot behaviors, setting the foundation for detailed formulation in subsequent sections of the paper.

\section{Learning from Demonstration framework for MRS}\label{method}

The core of our approach is encapsulated within a LfD framework, designed to enable MRS to learn and execute tasks through visual demonstration, as depicted in Fig.~\ref{fig:arch}. This process begins with the acquisition of task demonstrations \(D\) via a 2D camera within the vision system, capturing RGB imagery that provides the robots with visual cues necessary for task performance.

The \textbf{Vision Tracking Module} represents the first processing stage within our framework. It analyzes the video demonstrations, which vary in length depending on the complexity of the task at hand. The module's primary function is to extract prominent features concerning the robots \( R \) and objects \( O \) within each frame. These data are crucial to building a comprehensive understanding of the subtasks to be learned.

Following feature extraction, the \textbf{Task Policy Inference Module} takes over to dissect the demonstrations into smaller, interpretable segments. This process identifies \textit{Interaction Keypoints} \( IK \), which are critical for delineating subtasks and individual robot apriori skills from the demonstration. These keypoints enable the formation of a \textit{Task Policy} \( T_P \), a sequence of actions representing the learned skills and decisions.

Should the inference process reveal the necessity for a robot to learn a new contact-based skill not encapsulated within the existing repertoire, the \textbf{RL Module} is activated. Here, Soft Actor-Critic algorithms are employed to teach the robots these new skills, with the support of additional demonstrations as needed. The binary decision classifier \( C \) refines the reward structure, guiding the RL process to ensure effective skill acquisition.

The final step in our methodology involves the \textbf{Robot Execution Module}, which is tasked with the allocation of learned subtasks to the respective robots in the MRS. This module effectively translates the high-level \textit{Task Policy} \( T_P \) into actionable steps, using both preexisting apriori skills and newly learned RL policies for task execution. Through this vision-based LfD framework, we equip MRS with the capacity not only to replicate demonstrated tasks, but also to apply these learned behaviors to novel scenarios, bridging the gap between demonstration and autonomous execution.

\subsection{Vision Tracking Module} \label{vision}
The Vision Tracking Module stands as a critical component of our framework, underpinning both real-time task execution and subsequent learning. It tracks a range of environmental features—specifically, the 2D positions ($x$, $y$) of objects \( O \) and robots \( R \), as well as their goal positions \( G \). This module also captures more nuanced attributes such as object shape and color, charting the relationships among objects and robots.

The module assumes an occlusion-free environment, ensuring clear visibility of all objects and robots for reliable data capture. While this simplifies tracking, we anticipate future iterations to tackle partial occlusions, broadening the framework's versatility. The approach relies on the detectability of objects to maintain a finite, yet scalable, database for object profiles, supporting system adaptability by accommodating new objects as needed.

Object detection leverages the power of Mask R-CNN \cite{he2017mask}, a state-of-the-art deep learning technique known for its robust object detection and instance segmentation capabilities. Complementing this, shape and color detection are performed using functionalities provided by OpenCV, enabling the precise identification of object features such as color, shape, corners and centers upon detection. For robot tracking, we use ArUco markers \cite{lebedev2020accurate}, designed for efficient pose estimation.

\subsection{Task Policy Inference Module}\label{policy}

This module leverages visual features captured by the Vision Tracking Module to identify \textit{Interaction Keypoints}, which are critical for task segmentation and efficiency. Drawing inspiration from methods that utilize keypoints for task decomposition \cite{balakuntala2019extending,kannan2023upplied}, IKs enable the system to break complex tasks into smaller, manageable subtasks. This granularity allows each robot in the MRS to focus on specific segments of the task, streamlining the learning process, and improving system interpretability.

\noindent\textbf{Interaction Types and Definitions:} The module distinguishes four types of interactions within the MRS context:
\begin{enumerate}
    \item \textit{Object-Robot Interactions} (\(\phi_i\)): Interactions between objects and robots.
    \item \textit{Object-Object Interactions} (\(\psi_i\)): Interactions among objects.
    \item \textit{Robot-Robot Interactions} (\(\omega_i\)): Interactions among robots.
    \item \textit{Behavior Triggering Keypoints}: Environmental changes, such as the introduction of a new artifact, prompting a state transition from \(St\) to \(St^*\) and act as an action trigger.
\end{enumerate}

\noindent\textbf{Segmentation and Interaction Features:} \textit{Interaction Keypoints} aid in the temporal segmentation of tasks, highlighting each robot's role and interactions. Binary interaction features (\(\phi_i\), \(\psi_i\), \(\omega_i\)) denote the presence (1) or absence (0) of interactions. The system also tracks the relative and absolute positions of robots (\(Rr_i\), \(Ar_i\)) and objects (\(Ro_i\), \(Ao_i\)), along with motion flags \(f(Ro_i)\) and \(f(Rr_i)\) to indicate movement. We can denote each segment at time \(t\) for multiple robots as:
\begin{equation}
\Theta_{it} = (\theta_1, \theta_2, \ldots, \theta_r)
\end{equation} 
where task segment for a robot, \(\theta_r\), at an \textit{Interaction Keypoint} is defined as:
\begin{equation}
\theta_r = (\phi_i, Ro_i, f(Ro_i), Ao_i, \psi_i, \omega_i, Rr_i, f(Rr_i), Ar_i)
\end{equation}

These segments undergo classification via a decision tree \cite{charbuty2021classification,song2015decision} to assign class labels \(C_i\), corresponding to specific robot skills, whether pre-learned or acquired through RL.

\noindent\textbf{Policy Formation:} The resulting policy, \(\Pi\), sequences robot interactions and actions as follows:
\begin{equation}
\Pi = \{(C_1, g_1, \Theta_{1t}), (C_2, g_2, \Theta_{2t}), \ldots, (C_m, g_m, \Theta_{mt})\}
\end{equation}
where \(g_m\) denotes the goal state for each segment \(\Phi_{mt}\), reflecting the final environmental state within the segment. Class label \(C_m\) links to a specific skill set for execution during task realization.

\noindent\textbf{Task Dimensionality:} Task dimensionality, spanning from \(m\) to \(r \times m\), is adjusted based on the synchronization of the robot operations, ensuring the scalability of performance with the quantity of the robot, mainly affecting the data storage needs.

\subsection{RL Skill Learning} \label{rl}

Our framework utilizes RL within the LfD paradigm, specifically focusing on demonstrations to generate reward signals for agent training. This is crucial for introducing new, often complex, collaborative contact-based manipulation skills. Our model approaches these multi-robot manipulation challenges as model-free RL problems, activating this module when new skills are identified.

\noindent\textbf{Soft Actor-Critic (SAC) Method:} The SAC method is chosen for its efficiency and suitability for real-time execution within dynamic multi-robot systems. Combining actor-critic architecture with soft Q-learning, SAC ensures stable, adaptive learning, balancing exploration and exploitation. It employs a shared critic and individual actor networks for personalized learning, supported by a replay buffer for leveraging past experiences. This approach enhances responsiveness and learning diversity, following the SAC principles by \cite{haarnoja2018soft} and implemented in stable baselines \cite{raffin2021stable}.

\noindent\textbf{State Space and Action Space:} The SAC framework's state space is enriched with a 224x224 pixel image input processed via a Resnet 50 architecture, combined with precise robot and object positional data. Each robot's action space is defined by dual-speed parameters, controlling the motors of our mobile robot platform.

\noindent\textbf{Rewards:} Manual reward crafting for multi-robot tasks presents significant challenges, especially in achieving generalization across diverse tasks. By integrating additional demonstrations, our approach generalizes the reward function, leveraging a binary decision classifier for the determination of the reward signal, thus facilitating more intuitive and effective reward configurations.

Given the demonstrations' capability to highlight the task's goal through specific frames as shown in \cite{balakuntala2021learning}, we utilize these frames to discern positive and negative goal examples. Formally, let \(D = (I_n, y_n)\) represent the dataset, where \(I_n\) are the goal image frames, and \(y_n\) are the binary labels indicating positive (1) or negative (0) goal states.

The binary decision classifier \(C\) is trained on \(D\) to distinguish between goal states and non-goal states, optimizing its ability to function as a proxy for the reward function:
\begin{equation}
C(I_n) \rightarrow y_n
\end{equation}

The total reward signal \(R\) for the SAC is derived from the classifier's output and is structured as follows, incorporating weights for balance:
\begin{equation}
R = w_1 \cdot C(I_n) + w_2 \cdot IK_{reward} - w_3 \cdot IK Fail_{penalty}
\end{equation}
where $w_1$, $w_2$, and $w_3$ are weights that adjust the influence of each component on the total reward. \(IK_{reward}\) is designed to be minimal and a significant penalty for failures, emphasizing the achievement of the task's primary goal over mere interaction with keypoints. This nuanced reward strategy ensures that agents are incentivized to pursue the overarching objectives while maintaining focus over individual robot objectives efficiently.
\begin{figure}[t]
    \centering
    \includegraphics[width=0.40\textwidth]{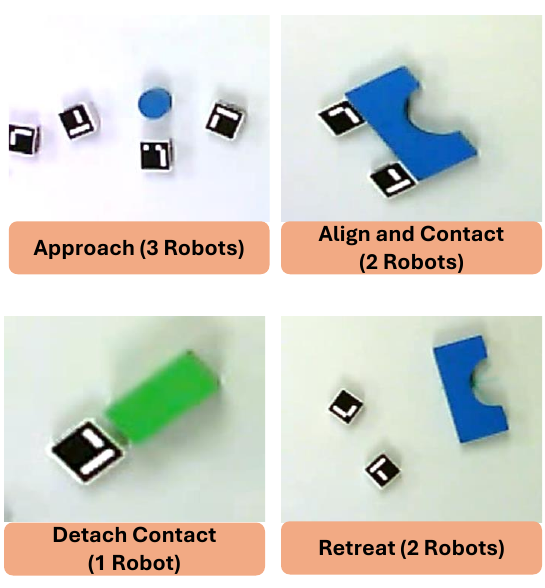}
    \caption{ Apriori skills are modeled as skills that each individual robot can perform. These are individual robot skills and do not constitute multi robot skills.}
    \label{fig:apriori}
       \vspace{-12pt}
 
\end{figure}

\subsection{Robot Execution Module}

The Robot Execution Module translates the \textit{Task Policy}, derived from Section \ref{policy}, into executable actions through Task Allocation. This allocation assigns robots either predefined apriori skills or RL skill policies from Section \ref{rl}.

Apriori skills are sensory-motor actions each robot executes independently, aligned with specific goal states within the policy. Our framework includes four critical apriori skills as depicted in Fig.~\ref{fig:apriori}: 1) \textit{Approach}, where a robot moves towards a target zone around an object or another robot; 2) \textit{Move to Contact}, achieving physical connection with an object; 3) \textit{Detach Contact}, where the robot disengages from the object; and 4) \textit{Retreat}, withdrawing from the object's vicinity. Executed with a precise low-level PD controller, these skills enable basic task performance and fundamental \textit{Interaction Keypoint} detection, such as object-object, object-robot and robot-robot interactions. This foundational capability is vital for understanding interactions within our system and also allows us to learn behavior-based tasks in a one-shot manner from a single demonstration.

\section{Experiments and Results}

\begin{figure}[h]

    \centering
    \includegraphics[width=0.35\textwidth,height = 4cm]{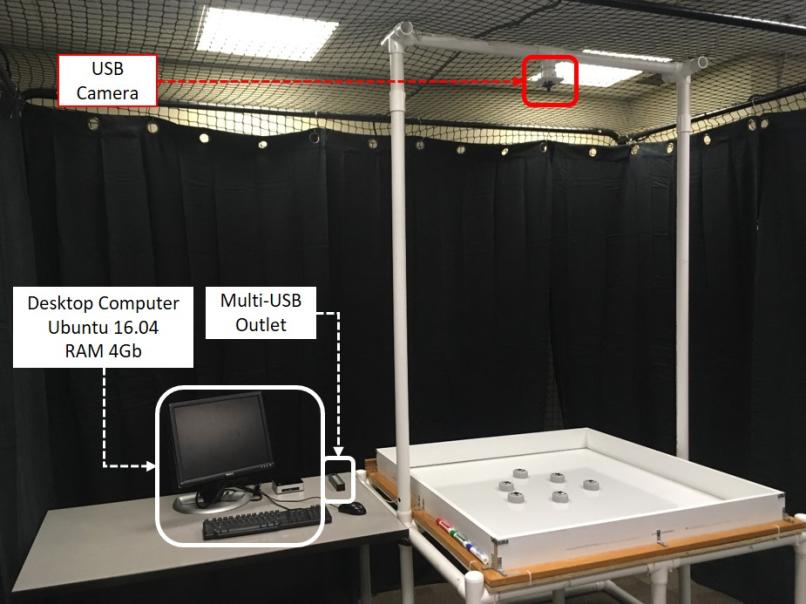}
    \caption{Experimental Testbed shows the Hamster robots in the environment with a mounted overhead camera attached to the system.}
    \label{fig:testbed}
    \vspace{-13pt}
\end{figure}

Our experiments were carried out on a test bed featuring a table with an overhead camera aligned parallel to the surface of the table. The Hamster mobile robots \cite{lee2021investigating} utilized for these experiments are equipped with dual-wheeled motors and IR proximity sensors, enabling precise navigation and interaction with their environment. This setup, depicted in Fig. \ref{fig:testbed}, is designed to facilitate both real-time training and the execution of tasks. Owing to the framework's ability to execute behavior-based tasks from a single demonstration, it was imperative to evaluate performance under real-world conditions. Consequently, we ensured that RL training for contact-based tasks was also conducted in the real environment. This approach circumvents the simulation-to-reality gap, which can often hinder the transition of contact-based manipulation skills to practical application. By maintaining consistency in training and testing exclusively in a real-world setting, we aimed to validate the framework's effectiveness in live scenarios.

\begin{figure*}[ht]
     \centering
     \begin{subfigure}{0.77\textwidth}
         \centering
         \includegraphics[width=\textwidth]{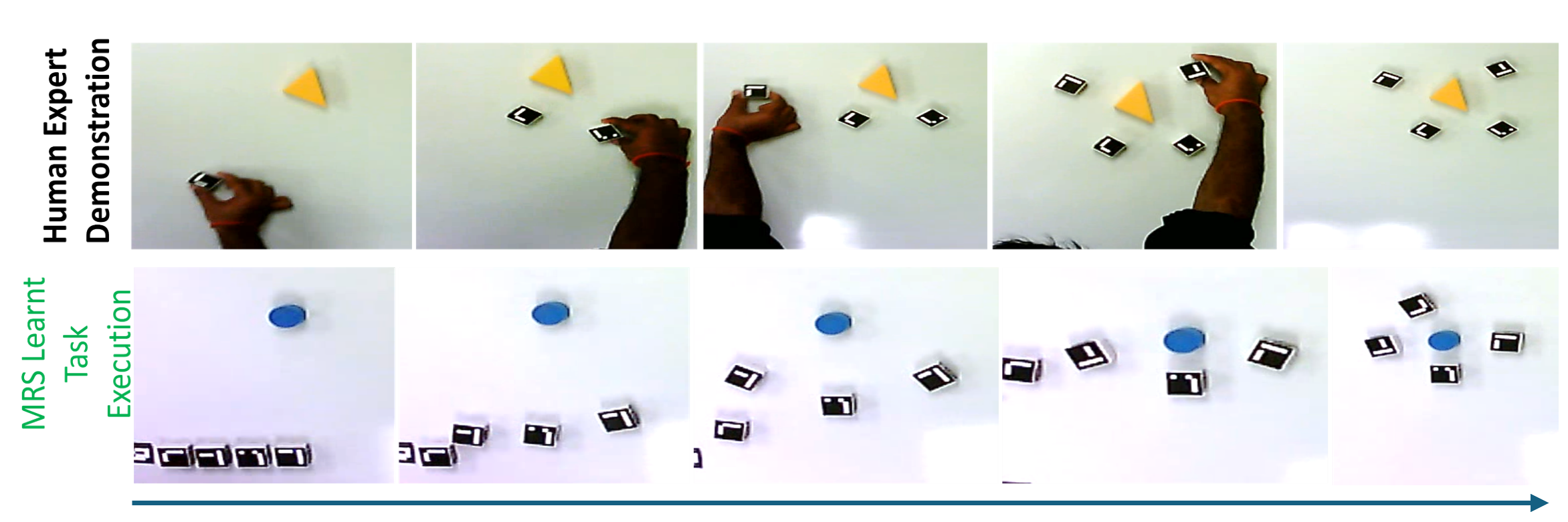}
         \vspace{-20pt}
         \caption{Demonstration and multi-robot execution of the \textbf{Intruder Attack} task.}
         \label{ResultIntrude}
     \end{subfigure}
     \hfill
     \begin{subfigure}{0.77\textwidth}
         \centering
         \includegraphics[width=\textwidth]{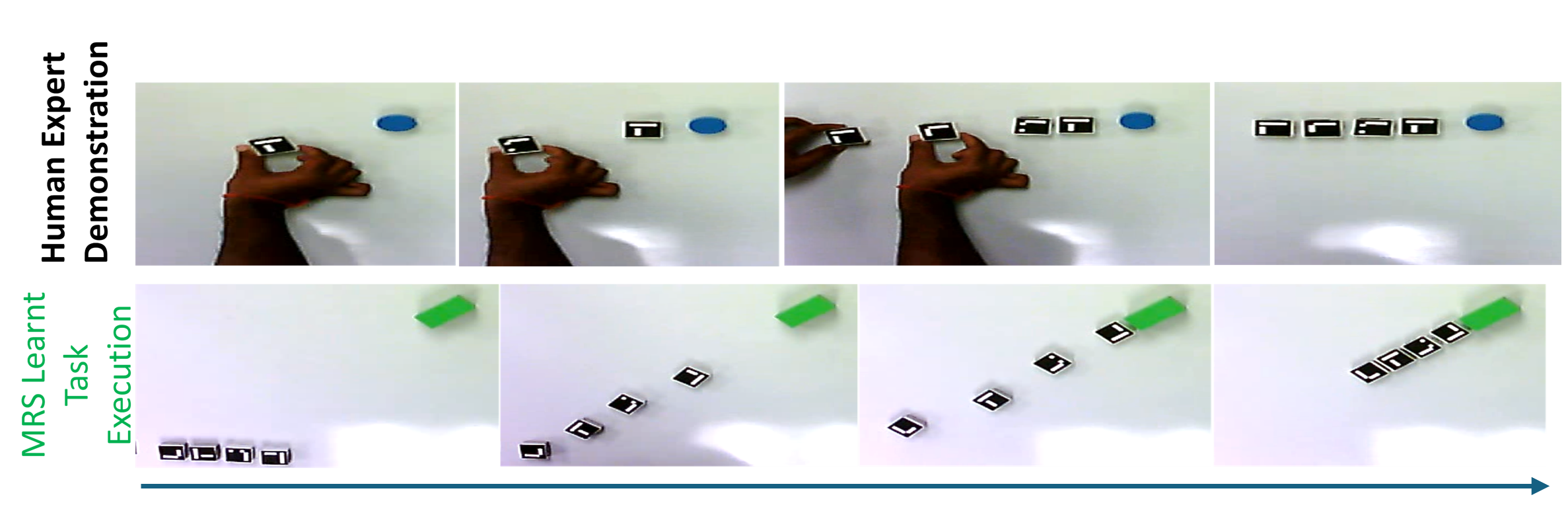}
         \vspace{-20pt}
      2   \caption{Demonstration and multi-robot execution of the \textbf{Leader Follower} task.}
         \label{ResultLine}
     \end{subfigure}
     \hfill
     \begin{subfigure}{0.77\textwidth}
         \centering
         \includegraphics[width=\textwidth]{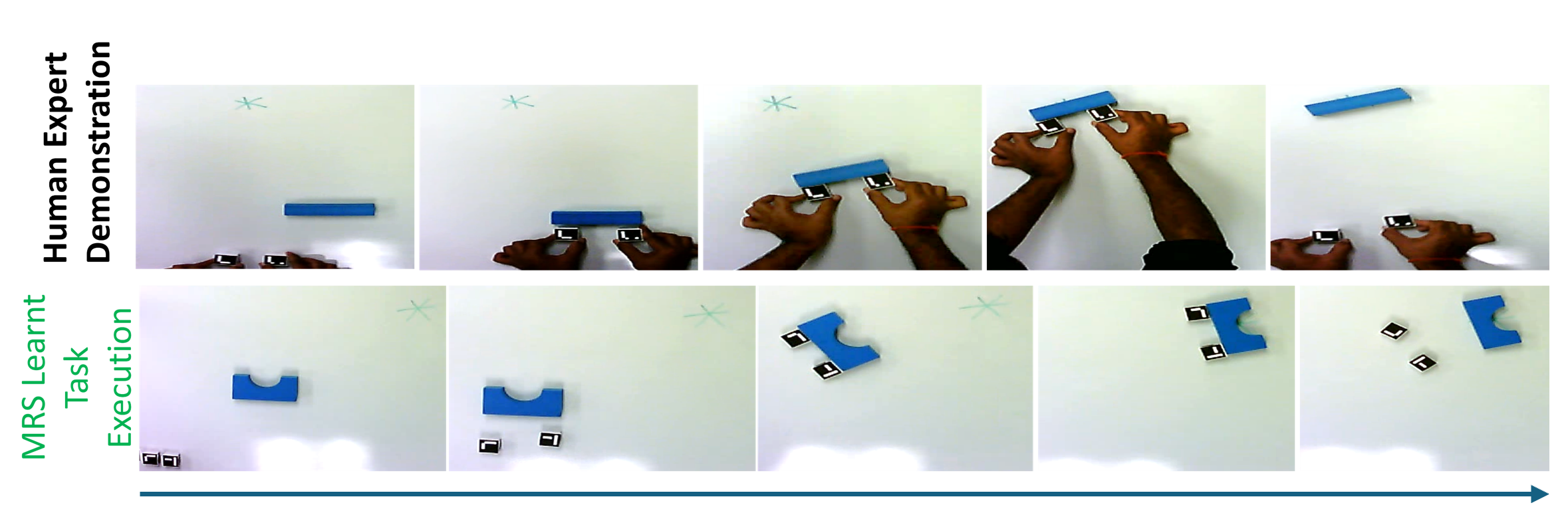}
         \vspace{-20pt}
         \caption{Demonstration and multi-robot execution of the \textbf{Object Transport} task.}
         \label{fig:ResultPush}
     \end{subfigure}
        \caption{Examples of demonstrations and task execution are presented. The objects used during demonstrations are different from the objects used during execution to showcase how our learning process is generalizable across different objects in the environment.  For example, in (a), the yellow object was used for demonstration, while the blue object was used for execution. Experiment videos showing more examples can be found in the supplementary video. }
        \label{fig:ResultVisual}

        \vspace{-12pt}
\end{figure*}

We evaluate the performance of our learning from demonstration for multi-robot systems framework across various tasks, including a) {Intruder Attack}, b) {Leader Follower}, c) {Object Transport}, d) {Object Rotate}, and e) {Object Color Sorting}. The tasks (a) and (b) involve \textit{behavior-based} objectives, while (c), (d) and (e) focus on \textit{contact-based} objectives. The behavior-based tasks require robots to demonstrate specific behaviors, while the contact-based tasks involve rich manipulation, particularly in \textit{Object Transport}, a well-studied area in multi-robot domains. Additionally, we implement a baseline method, a naive RL approach using the SAC algorithm. In this method, rewards are manually engineered according to each task's nature.

The principal metric for assessment was the success rate (SR), which represents the proportion of successful trials within a set of 30. Success was strictly defined by the robots' ability to fulfill the task's requirements, be it encircling an intruder with a defined boundary or executing accurate color-specific object sorting. 

It is worth to note that we primarily focus on measuring the success or failure of task completion, employing metrics such as SR. This choice is influenced by the current state of Multi-Robot LfD approaches and the unique challenges they present. Present-day multi-robot LfD approaches often lack sophistication, and achieving reasonable skill goals is a significant accomplishment. Thus, assessing the success or failure of these tasks provides meaningful insights into the framework's capabilities. The absence of standardized evaluation parameters or approaches in LfD is due to the highly task-specific nature of these frameworks. Unlike more established fields, there is no one-size-fits-all approach or a baseline for comparison. Another challenge is the diversity of robot platforms used in LfD, each with varying physical structures, characteristics, and constraints. This diversity makes it challenging to conduct direct quantitative comparisons between different LfD approaches.

Given these factors, our emphasis on success or failure in task completion, along with the inclusion of a baseline RL method, provides a practical and relevant evaluation strategy for our framework.

\subsection{Intruder Attack}
The \textit{Intruder Attack} task required a team of robots to encircle an intruder. Given only one expert demonstration, the robots had to adapt to variations in the intruder's features and positions. This behavior-oriented task leveraged the robots' apriori skills, specifically designed for approach and surround strategies. The reward function of the naive RL approach was predicated on the proximity of each robot to the intruder, aiming for a formation within 10 $cm$ of the target. Success was evaluated based on the final positions of the robots in relation to the intruder and each other, the experiment achieving a success rate of 95\% in a task with 3 robots and 92\% with 5 robots when provided with just single demonstration, despite variations in environment and configurations.

\subsection{Leader Follower}
The \textit{Leader Follower} task involved creating a follower formation behind a moving leader object, using just one expert demonstration as a reference. This behavior-oriented task utilized the robots' innate apriori skills to autonomously align in a sequential formation. The naive RL method calculated rewards based on the distance to maintain a coherent line behind the leader. The successful creation of this formation, as depicted in the accompanying figure, was achieved with a success rate 100\% when provided with a single demonstration, which underscores the effectiveness of the behaviors demonstrated.

\subsection{Object Transport}
For the \textit{Object Transport} task, pairs of robots were required to collaboratively move an object to a designated target location. The task complexity necessitated 50 demonstrations to effectively train the classifier for a nuanced reward signal. Employing a SAC architecture with a two-layer Multilayer Perceptron (MLP) network of 64 units each, the robots were trained over an 8-hour period. The SR, determined by the final proximity of the object to the goal, was recorded at 80\%, demonstrating the robustness of the framework in facilitating cooperative transport.

\subsection{Object Rotate}
The \textit{Object Rotate} task demanded precision as robots worked together to rotate an object by 180 degrees. To accommodate the required precision, the task also involved 50 demonstrations to refine the classifier's reward mechanism. After 6 hours of training with a SAC model similar to the \textit{Object Transport} task, the robots accomplished a 67\% SR, with a permissible margin of error of +/- 2 degrees, highlighting our framework's potential in tasks requiring exact maneuvers.

\subsection{Object Color Sorting}
Building on the \textit{Object Transport} mechanics, the \textit{Object Color Sorting} task required the robots to sort objects by color into corresponding goals. No additional RL training was required, as the task extended previously learned transport skills. With a batch of 4 objects of varying colors introduced sequentially, the robots achieved a 77\% SR, based on the accurate sorting of all objects. Errors were mainly attributed to the transport phase rather than the sorting, as robots consistently recognized the correct location of the goal.

The success rates, as illustrated in Table \ref{table:1}, validate the effectiveness of our framework in both behavior-based and contact-based tasks. The high SR across different task types signifies our framework's competency in real-time execution and adaptability to task variations, especially when compared to traditional training intensive methods such as the baseline RL.

\begin{table}
\centering
\vspace{-0pt}

\caption{Success rates for all tasks. Each task underwent 30 trials.}
\label{table:1}
\begin{tabular}{cccc} 
\hline
\textbf{Task Type} & \textbf{Task} & \begin{tabular}[c]{@{}c@{}}\textbf{Proposed}\\\textbf{ Methods (\%)}\end{tabular} & \textbf{Baseline (\%)} \\ 
\hline
\multirow{4}{*}{\begin{tabular}[c]{@{}c@{}}Behaviour-\\based \\Tasks\end{tabular}} & \begin{tabular}[c]{@{}c@{}}Intruder Attack\\(3 Robots)\end{tabular} & 95 & 60 \\
 & \begin{tabular}[c]{@{}c@{}}Intruder Attack\\(5 Robots)\end{tabular} & 92 & 38 \\
 & \begin{tabular}[c]{@{}c@{}}Leader Follower\\(3 Robots)\end{tabular} & 100 & 65 \\
 & \begin{tabular}[c]{@{}c@{}}Leader Follower\\(4 Robots)\end{tabular} & 100 & 58 \\ 
\hline
\multirow{3}{*}{\begin{tabular}[c]{@{}c@{}}Contact-\\based \\Tasks\end{tabular}} & Object Transport & 80 & 40 \\
 & Object Rotate & 67 & 42 \\
 & Object Sorting & 88 & 15 \\
\hline
\end{tabular}
\vspace{-15pt}

\end{table}

Furthermore, an ablation study on contact-based tasks, presented in Table II, examined the performance under three conditions: using the full proposed method, only IK rewards, and only Classifier rewards. The complete framework consistently outperformed the other conditions, reinforcing the synergy between the Interaction Keypoints and the classifier-based rewards in our method. For object transport, the success rates dropped to 47\% and 62\% when relying solely on IK rewards and Classifier rewards, respectively. Object Rotate saw similar trends, with success rates of 24\% for IK rewards and 48\% for Classifier rewards, confirming the integral role of our hybrid reward strategy.

\begin{table}
\centering
\caption{Success rates for all contact-based tasks with the skills learned under different reward conditions. Each task underwent 30 trials.}
\vspace{-5pt}

\label{table:2}
\begin{tblr}{
  width = \linewidth,
  colspec = {Q[256]Q[210]Q[212]Q[244]},
  cells = {c},
  hline{1-2,4} = {-}{},
}
\textbf{Task}    & {\textbf{Proposed}\\\textbf{ Methods (\%)}} & {\textbf{Only IK}\\\textbf{ Rewards (\%)}} & {\textbf{Only Classifier}\\\textbf{ Rewards (\%)}} \\
Object Transport & 80                                          & 47                                         & 62                                                 \\
Object Rotate    & 67                                          & 24                                         & 48                                                 
\end{tblr}
\vspace{-5pt}

\end{table}

These results collectively emphasize our framework's efficiency and its ability to generalize across various MRS tasks with fewer demonstrations needed, paving the way for practical applications in dynamic environments.

\section{Discussion}
Our investigation into the learning from demonstration  framework for multi-robot systems has yielded promising results, revealing the framework's potential in a real-time, real-world setting. While the current version demonstrates a robust capability for managing tasks defined by discrete interaction keypoints, there is room to extend this proficiency to trajectory-based tasks, thereby broadening the scope of the framework's applicability.

One significant realization from our experiments is the value of real-time training and testing. By deploying our framework directly in the real-world environment, we have effectively sidestepped the simulation-to-reality (sim2real) gap that often plagues contact-based manipulation tasks. However, scalability remains a key challenge. The complexity of RL training increases with the number of robots, and thus the design of strategies for scalable, task-agnostic learning is an important direction for future research. This could involve developing more efficient training algorithms or creating frameworks that allow for incremental learning as new robots are added.

Another exciting prospect is robot domain transfer, which involves applying a universal LfD framework to various robot types, enabling skill transfer across different platforms and simplifying the deployment of MRS. Moreover, the domain of heterogeneous multi-robot tasks, where different robot types like aerial drones and ground robots collaborate, presents a fertile ground for innovation. Exploring our framework’s flexibility in such mixed-robot environments is an important next step.

Although this paper represents an initial step toward realizing a generalized LfD framework capable of handling a multitude of tasks in various environments, the journey ahead is expansive. Continuous research efforts are crucial to overcome existing limitations and to harness the full potential of MRS. The success we have demonstrated in a real-world context lays a strong foundation for future endeavors, aspiring towards a universally adaptable MRS framework.

\section{Conclusion}
This paper presents an innovative Learning from Demonstration (LfD) framework for Multi-Robot Systems (MRS), leveraging visual demonstrations and a binary decision classifier to streamline skill acquisition and task execution. By reducing the need for extensive demonstrations, our approach addresses the challenge of data intensity in LfD research. Showing great promise for robust multi-robot learning, we plan to enhance our framework's scalability, support for heterogeneous robot teams, and trajectory-based skill inclusion. These advancements pave the way for a future of highly efficient and adaptable robot learning and collaboration, advancing autonomous systems.

\bibliographystyle{IEEEtran}
\balance
\bibliography{refs_MRS}
\end{document}